# The Computation of All 4R Serial Spherical Wrists With an Isotropic Architecture

Damien Chablat[*] and Jorge Angeles[**]

[*]Institut de Recherche en Communications et Cybernétique de Nantes[1]

1, rue de la Noë, 44321 Nantes, France

[**]Department of Mechanical Engineering & Centre for Intelligent Machines, McGill University

817 Sherbrooke Street West, Montreal, Canada H3A 2K6

Damien.Chablat@irccyn.ec-nantes.fr, angeles@cim.mcgill.ca

Phone: 33 2 40 37 69 54 - Fax: 33 2 40 37 69 30

## 1   Abstract

A spherical wrist of the serial type with $n$ revolute (R) joints is said to be isotropic if it can attain a posture whereby the singular values of its Jacobian matrix are all equal to $\sqrt{n/3}$. What isotropy brings about is robustness to manufacturing, assembly, and measurement errors, thereby guaranteeing a maximum orientation accuracy. In this paper we investigate the existence of redundant isotropic architectures, which should add to the dexterity of the wrist under design by virtue of its extra degree of freedom. The problem formulation, for $n = 4$, leads to a system of eight quadratic equations with eight unknowns. The Bezout number of this system is thus $2^8 = 256$, its BKK bound being 192. However, the actual number of solutions is shown to be 32. We list all solutions of the foregoing algebraic problem. All these solutions are real, but distinct solutions do not necessarily lead to distinct manipulators. Upon discarding those algebraic solutions that yield no new wrists, we end up with exactly eight distinct architectures, the eight corresponding manipulators being displayed at their isotropic postures.

## 2   Introduction

The kinematic design of redundant spherical wrists under isotropy conditions is the subject of this paper. A manipulator is called *isotropic* if its Jacobian matrix can attain isotropic values at certain postures [1,6]. A matrix, in turn, is called *isotropic* if its singular values are all identical and nonzero. Furthermore, the matrix condition number can be defined as the ratio of its greatest to its smallest singular values [2]. Thus, isotropic matrices have a minimum condition number of unity. The kinematic structure of industrial manipulators are frequently decoupled into a positioning and an orientation submanipulator. The latter is designed with revolute joints whose axes intersect. However, when these three joints are coplanar, the manipulator becomes singular [3]. As a means to cope with singularities, redundant wrists have been suggested [4]. An extensive bibliography on the design of spherical wrists can be found in [5-6].

---

An isotropic Jacobian matrix is of interest because its condition number is unity, and hence, a minimum [2,6]. It is noteworthy that the condition number gives an upper bound for the relative roundoff-error amplification, upon solving a system of linear equations, with respect to the relative roundoff-error in the data. The latter are contained in the numerical entries of the Jacobian matrix and in the twist [7]. In fact, machining tolerances and assembly errors bring about additional errors in the Jacobian entries.

Prior to our analysis leading to the architectures sought, we recall a few geometric concepts in the subsection below.

### 2.1    *Isotropic Sets of Points on the Unit Sphere*

Consider the set $S \equiv \{P_k\}_1^n$ of $n \geq 3$ points on the unit sphere, of position vectors $\{\mathbf{e}_k\}_1^n$. Apparently, all the vectors of the foregoing set are of unit Euclidean norm. The *second-moment tensor* $\mathbf{H}$ of $S$ is defined as

$$\mathbf{H} = \sum_1^n \mathbf{e}_k \mathbf{e}_k^T \tag{1}$$

The set $S$ is said to be *isotropic* if its second-moment tensor is isotropic. Since $\mathbf{H}$ is symmetric and positive-definite, it is isotropic if it is proportional to the $3 \times 3$ identity matrix $\mathbf{1}$, the proportionality factor, denoted here with $\sigma^2$, being the square of the triple singular value of $\mathbf{H}$. In our case, apparently, the singular values of $\mathbf{H}$ coincide with its eigenvalues.

We note that, if $S$ is the set of vertices of a Platonic solid, then $\mathbf{H}$ is isotropic. Table 1 records the values of $n$ and $\sigma$ for each Platonic solid.

|   | Tetrahedron | Cube | Octahedron | Dodecahedron | Icosahedron |
|---|---|---|---|---|---|
| n | 4 | 8 | 6 | 20 | 12 |
| σ | 4/3 | 8/3 | 2 | 20/3 | 4 |

**Table 1: The values of *n* and  σ  for the Platonic solids**

**Remark 1**: It is apparent that, if a point $P_k$ of an arbitrary set $S$ of points on the unit sphere is replaced by its *antipodal* $Q_k$, of position vector $\mathbf{q}_k = -\mathbf{e}_k$, then the second-moment tensor $\mathbf{H}$ of $S$ is *preserved*.

The replacement of a point on the unit sphere by its antipodal will be termed, henceforth, *antipodal exchange*. As a consequence of Remark 1, then, the isotropy of a set of points on the unit sphere is preserved under any antipodal exchanges.

## 3    Isotropic Sets of Points on the Unit Sphere

The basic sets of isotropic points are thus the sets of vertices of the Platonic solids. Hence,

**Definition** (Fundamental isotropic set): An isotropic set $S$ of $n$ points on the unit sphere is called *fundamental* if its elements are the vertices of a Platonic solid (inscribed, of course, on the unit sphere.)

It is noteworthy that isotropic sets on the unit sphere exist that are none of the Platonic solids, e.g., the 64-vertex polyhedron defined by the molecule of the buckminsterfullerene, popularly known as the Buckyball [8]. The name comes from the architect R. Buckminster Fuller, who used this polyhedron as the structure of the geodesic dome built on occasion of the Universal Exhibit of 1967 in Montreal. This polyhedron is also present in the patterns of soccer balls.

Also note that the set of points on the unit sphere leading to an isotropic architecture for a spherical wrist need not be laid out with the center of the sphere as its centroid, which is a condition found for points in the plane [9]. For example, the three points of intersection of the unit sphere with the axes of an orthogonal coordinate frame with origin at the center of the sphere define an isotropic spherical wrist, namely, the one most commonly encountered in commercial manipulators, yet the above set of points corresponds to none of the Platonic solids. This three-revolute wrist is termed *orthogonal* because its neighboring axes make right angles.

Fundamental sets of isotropic points are important because they allow the derivation of new sets by simple operations, as described below.

### 3.1   Properties of Isotropic Sets of Points on the Unit Sphere

First, note

**Lemma 1:** An isotropic set $S$ of points on the unit sphere remains isotropic under any *isometric transformation* of the set.

An isometry being either a rigid-body rotation or a reflection, the foregoing lemma should be obvious. Moreover, rigid-body rotations of isotropic sets are uninteresting because they amount to looking at the given set from a different viewpoint. However, distinct isotropic sets can be derived from reflections of isotropic sets about planes or lines. Nevertheless, as shown in the Appendix, a reflection about a line amounts to a rigid-body rotation about the line through an angle of $\pi$. As a consequence, then, only reflections about planes will be considered when defining new isotropic sets from fundamental ones.

### 3.2   Derived Isotropic Sets of Points

We show in this subsection, with a numerical example, how new sets of isotropic points on the unit sphere can be derived from a fundamental set by application of reflections about planes. Now, since the reflection plane can be defined in infinitely-many ways, a correspondingly infinite number of reflections is possible. We are interested only in *linearly-independent* reflections, which are, apparently, only three, one about each of three mutually orthogonal planes.

A fundamental isotropic set of four points, namely, the vertices of a regular tetrahedron inscribed in the unit sphere, is given below:

$$\mathbf{e}_1 = \begin{bmatrix} 1 \\ 0 \\ 0 \end{bmatrix}, \; \mathbf{e}_2 = \begin{bmatrix} -1/3 \\ -2\sqrt{2}/3 \\ 0 \end{bmatrix}, \; \mathbf{e}_3 = \begin{bmatrix} -1/3 \\ \sqrt{2}/3 \\ \sqrt{6}/3 \end{bmatrix}, \; \mathbf{e}_4 = \begin{bmatrix} -1/3 \\ \sqrt{2}/3 \\ -\sqrt{6}/3 \end{bmatrix} \qquad (2a)$$

We produce now three new sets of isotropic points by reflecting the foregoing set onto the three coordinate planes, *y-z*, *x-z* and *x-y*, successively, *i.e.*,

•  The reflection with respect to the *y-z* plane gives

$$\mathbf{e}_1' = \begin{bmatrix} -1 \\ 0 \\ 0 \end{bmatrix}, \ \mathbf{e}_2' = \begin{bmatrix} 1/3 \\ -2\sqrt{2}/3 \\ 0 \end{bmatrix}, \ \mathbf{e}_3' = \begin{bmatrix} 1/3 \\ \sqrt{2}/3 \\ \sqrt{6}/3 \end{bmatrix}, \ \mathbf{e}_4' = \begin{bmatrix} 1/3 \\ \sqrt{2}/3 \\ -\sqrt{6}/3 \end{bmatrix} \tag{2b}$$

- The reflection with respect to the *x-z* plane gives

$$\mathbf{e}_1'' = \begin{bmatrix} 1 \\ 0 \\ 0 \end{bmatrix}, \ \mathbf{e}_2'' = \begin{bmatrix} -1/3 \\ 2\sqrt{2}/3 \\ 0 \end{bmatrix}, \ \mathbf{e}_3'' = \begin{bmatrix} -1/3 \\ -\sqrt{2}/3 \\ \sqrt{6}/3 \end{bmatrix}, \ \mathbf{e}_4'' = \begin{bmatrix} -1/3 \\ -\sqrt{2}/3 \\ -\sqrt{6}/3 \end{bmatrix} \tag{2c}$$

- The reflection with respect to the *x-y* plane gives

$$\mathbf{e}_1''' = \begin{bmatrix} 1 \\ 0 \\ 0 \end{bmatrix}, \ \mathbf{e}_2''' = \begin{bmatrix} -1/3 \\ -2\sqrt{2}/3 \\ 0 \end{bmatrix}, \ \mathbf{e}_3''' = \begin{bmatrix} -1/3 \\ \sqrt{2}/3 \\ -\sqrt{6}/3 \end{bmatrix}, \ \mathbf{e}_4''' = \begin{bmatrix} -1/3 \\ \sqrt{2}/3 \\ \sqrt{6}/3 \end{bmatrix} \tag{2d}$$

## 4   Isotropic Spherical Wrists

Most serial wrists encountered in manipulators are provided with three revolute joints. We start with a general *n*-revolute spherical wrist, as depicted in Fig. 1, with Jacobian matrix **J** given by [7]

$$\mathbf{J} = \begin{bmatrix} \mathbf{e}_1 & \mathbf{e}_2 & \cdots & \mathbf{e}_n \end{bmatrix}^T \tag{3}$$

where, we recall, $\mathbf{e}_k$ is the unit vector indicating the direction of the *k*th revolute axis. We display below the kinematic relation between the joint-rate vector $\dot{\boldsymbol{\theta}}$ and the angular-velocity vector $\boldsymbol{\omega}$ of the end-effector (EE):

$$\mathbf{J}\dot{\boldsymbol{\theta}} = \boldsymbol{\omega} \tag{4}$$

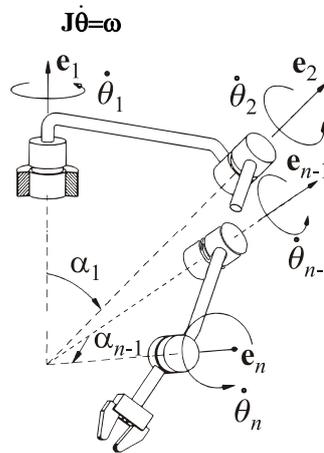

**Figure 1 A general *n*-revolute spherical wrist**

It should be apparent that

**Remark 2:** A set $\{\mathbf{e}_k\}_1^n$ of unit vectors produces *n*! Jacobian matrices, and hence, *n*! *distinct wrists*.

Kinetostatic isotropy with regard to force and motion transmission [7] requires that the singular values of the Jacobian matrix be all identical and nonzero, *i.e.*,

$$\mathbf{J}\mathbf{J}^T = \sigma^2 \mathbf{1} \tag{5}$$

where $\sigma$ is the common singular value, of multiplicity three, and $\mathbf{1}$ is, as defined earlier, the $3 \times 3$ identity matrix. The isotropy condition thus leads to

$$\sum_{1}^{n} \mathbf{e}_k \mathbf{e}_k^T = \sigma^2 \mathbf{1} \tag{6}$$

The value of $\sigma$ is found by taking the trace of both sides of eq.(6), which yields

$$\sum_{1}^{n} \mathbf{e}_k \cdot \mathbf{e}_k = 3\sigma^2 \tag{7}$$

and hence,

$$\sigma = \sqrt{\frac{n}{3}} \tag{8}$$

*i.e.*, if $\mathbf{J}$ is isotropic, then (a) every pair of $n$-dimensional rows of $\mathbf{J}$ is orthogonal and (b) the three rows of $\mathbf{J}$ have the same Euclidean norm, namely, $\sqrt{n/3}$.

Now, since it does not appear practical to design wrists with more than four revolutes, we limit ourselves, in the balance of the paper, to four-revolute manipulators, *i.e.*, we set $n = 4$ in the balance of the paper. Note that a serial spherical wrist with five or more revolutes would make it extremely heavy for the first actuator. Not only this; link interference is a problem that grows with the number of joints, and hence, of links.

## 5    Four-Axis Isotropic Spherical Wrists

In this section we obtain all possible four-revolute serial spherical wrists with isotropic architectures. The algebraic problem at hand consists in finding the set of vectors $\{\mathbf{e}_k\}_1^4$ associated with the set of points $\{P_k\}_1^4$ that verify the isotropy conditions of eq.(6). Without loss of generality, we define $\mathbf{e}_1$ parallel to the *x*-axis of the coordinate frame at hand; then, we let $\mathbf{e}_2$ lie in the *y-z* plane of the same frame, while the remaining two vectors are left arbitrary. We thus have

$$\mathbf{e}_1 = \begin{bmatrix} 1 \\ 0 \\ 0 \end{bmatrix}, \ \mathbf{e}_2 = \begin{bmatrix} c \\ s \\ 0 \end{bmatrix}, \ \mathbf{e}_3 = \begin{bmatrix} x \\ y \\ z \end{bmatrix}, \ \mathbf{e}_4 = \begin{bmatrix} u \\ v \\ w \end{bmatrix} \tag{9}$$

in which $c \equiv \cos(\alpha_1)$ and $s \equiv \sin(\alpha_1)$, as per Fig. 1. The isotropy condition (6), in terms of the foregoing components and with $\sigma^2 = 4/3$, yields, then,

$$1 + c^2 + x^2 + u^2 = 4/3 \tag{10a}$$

$$s^2 + y^2 + v^2 = 4/3 \tag{10b}$$

$$z^2 + w^2 = 4/3 \tag{10c}$$

$$c\,s + x\,y + u\,v = 0 \tag{10d}$$

$$z\,y + w\,v = 0 \tag{10e}$$

$$x z + u w = 0 \tag{10f}$$

Moreover, we have the normality of $\mathbf{e}_2$ and $\mathbf{e}_3$:

$$c^2 + s^2 = 1 \tag{10g}$$

$$x^2 + y^2 + z^2 = 1 \tag{10h}$$

the normality of $\mathbf{e}_4$ being embedded in the foregoing system of equations. Indeed, this is obtained upon adding eqs.(10a-c) and subtracting this sum from the sum of eqs.(10g & h). We have now eight quadratic equations for eight unknowns. The Bezout number of this system is thus $2^8 = 256$, which means that up to 256 solutions are to be expected, including real and complex, as well as multiple solutions. Moreover, the BKK bound [10-11] of the same system, known to give lowest bound, turns out to be 192. It will be shown presently that this number is too big, the total number of solutions being substantially smaller. In order to find the solutions of interest, we eliminate successively all the unknowns but $u$ to obtain a monovariate polynomial in this unknown. First, we solve for $x$, $y$ and $c$ from eqs.(10d-f), thus obtaining

$$x = -\frac{wu}{z} \tag{11a}$$

$$y = -\frac{vw}{z} \tag{11b}$$

$$c = \frac{u(w\,y\,-\,v\,z)}{s\,z} \tag{11c}$$

Upon substituting eqs.(11a-c) into eqs.(10a-c) and (10g & b), we obtain

$$w^2\,u^2\,+\,v^2\,w^2\,+\,z^4\,-\,z^2\,=\,0 \tag{12a}$$

$$3s^2z^2 + 3v^2w^2 + 3v^2z^2 - 4z^2 = 0 \tag{12b}$$

$$3z^2 + 3w^2 - 4 = 0 \tag{12e}$$

$$3u^2w^4v^2 + 6u^2w^2v^2z^2 + 3u^2v^2z^4 + 3w^2u^2s^2z^2 + 3u^2z^4s^2 - s^2z^4 = 0 \tag{12c}$$

$$u^2w^4v^2 + 2u^2w^2v^2z^2 + u^2v^2z^4 + s^4z^4 - s^2z^4 = 0 \tag{12d}$$

It is noteworthy that the system of eqs.(12a-e) contains only second and fourth powers of all the unknowns, which allows for a recursive solution, as we shall show below. First, from eq.(12e) we solve for $w$:

$$w = \pm\frac{1}{3}\sqrt{12 - 9z^2} \tag{13a}$$

Upon substitution of eq.(13a) into eq.(12b) we obtain $s$:

$$s = \pm\frac{2}{3}\frac{\sqrt{3z^2 - 3v^2}}{z} \tag{13b}$$

Likewise, substitution of eq.(13b) into eq.(12a) yields $v$:

$$v = \pm\sqrt{z^2 - 4u^2} \qquad (13c)$$

Finally, substitution of eq.(13c) into eq.(12d) leads to $z$:

$$z = \pm\sqrt{6}u \qquad (13d)$$

Now, substitution of eqs.(13a-d) into eq.(12c) leads to a monovariate polynomial:

$$u \, (3u - 1)(3u + 1) = 0 \qquad (13e)$$

The above equation reduces, in fact, to a quadratic equation because $u$ cannot vanish, as we shall show presently. Thus, the two possible solutions for eq.(13e) are

$$u = \pm\frac{1}{3} \qquad (14)$$

We thus have a set of five quadratic expressions for the five unknowns $u$, $z$, $v$, $s$ and $w$, which means that we have found $2^5 = 32$ distinct solutions, as displayed in Table 2. Therefore, the Bezout number of this system overestimates the number of solutions by a factor of eight, while the BKK bound by a factor of six.

The solutions can now be readily computed recursively. Indeed, eqs.(13a-c) lead to

$$v = \pm\sqrt{2}u \qquad (15a)$$

$$s = \pm\frac{2\sqrt{2}}{3} \qquad (15b)$$

$$w = \pm\frac{1}{3}\sqrt{6(2 - 9u^2)} \qquad (15c)$$

the remaining unknowns being computed from eqs.(13d), (11a), (11b) and (11c), in this order.

Now we show that none of the unknowns can vanish. We do this by noting that:

1.  If $u = 0$, then $v = 0$ by virtue of eq.(15a), while $w = \pm 2\sqrt{2}/3$, by virtue of eq.(15c), thereby violating the normality condition on $\mathbf{e}_4$. Also note that $u = 0$ leads to $c = 0$ by virtue of eq.(11c), but this value, along with that of $s$ given by eq.(15b), violates the normality condition (10g);

2.  If $v = 0$, then $u = 0$ by virtue of eq.(15a) but, according to item 1, this is impossible;

3.  If $w = 0$, then, by virtue of eqs.(15a & b), $x = y = 0$; additionally, by virtue of eqs.(13d) and (14), $z = \pm\sqrt{6}/3$, thereby violating the normality condition (10h);

4.  If $x = 0$, then, according to eq.(13d), either $u = 0$ or $w = 0$, but none of these can vanish, according to 1 and 3;

5.  If $y = 0$, then, by virtue of eq.(11b) either $v = 0$ or $w = 0$, but these alternatives are not plausible according to items 2 and 3;

6.  If $z = 0$, then $u = 0$, according with eq.(13d); however, by virtue of item 1, this is not possible;

7.  If $c = 0$, then the normality of $\mathbf{e}_2$ requires that $s = \pm 1$, but this is impossible by virtue of eq.(15b);

8.  From eq.(15b), $s$ cannot vanish, thereby showing that none of the unknowns can vanish.

| Solution # | $c$ | $s$ | $x$ | $y$ | $z$ | $u$ | $v$ | $w$ |
|---|---|---|---|---|---|---|---|---|
| 1 | $1/3$ | $-2\sqrt{2}/3$ | $-1/3$ | $-2\sqrt{2}/3$ | $\sqrt{6}/3$ | $1/3$ | $2\sqrt{2}/3$ | $\sqrt{6}/3$ |
| 2 | $1/3$ | $-2\sqrt{2}/3$ | $-1/3$ | $-2\sqrt{2}/3$ | $\sqrt{6}/3$ | $-1/3$ | $-2\sqrt{2}/3$ | $-\sqrt{6}/3$ |
| 3 | $1/3$ | $-2\sqrt{2}/3$ | $-1/3$ | $-2\sqrt{2}/3$ | $-\sqrt{6}/3$ | $1/3$ | $2\sqrt{2}/3$ | $-\sqrt{6}/3$ |
| 4 | $1/3$ | $-2\sqrt{2}/3$ | $-1/3$ | $-2\sqrt{2}/3$ | $-\sqrt{6}/3$ | $-1/3$ | $-2\sqrt{2}/3$ | $\sqrt{6}/3$ |
| 5 | $1/3$ | $-2\sqrt{2}/3$ | $1/3$ | $2\sqrt{2}/3$ | $\sqrt{6}/3$ | $1/3$ | $2\sqrt{2}/3$ | $-\sqrt{6}/3$ |
| 6 | $1/3$ | $-2\sqrt{2}/3$ | $1/3$ | $2\sqrt{2}/3$ | $\sqrt{6}/3$ | $-1/3$ | $-2\sqrt{2}/3$ | $\sqrt{6}/3$ |
| 7 | $1/3$ | $-2\sqrt{2}/3$ | $1/3$ | $2\sqrt{2}/3$ | $-\sqrt{6}/3$ | $1/3$ | $2\sqrt{2}/3$ | $\sqrt{6}/3$ |
| 8 | $1/3$ | $-2\sqrt{2}/3$ | $1/3$ | $2\sqrt{2}/3$ | $-\sqrt{6}/3$ | $-1/3$ | $-2\sqrt{2}/3$ | $-\sqrt{6}/3$ |
| 9 | $1/3$ | $2\sqrt{2}/3$ | $-1/3$ | $2\sqrt{2}/3$ | $\sqrt{6}/3$ | $1/3$ | $-2\sqrt{2}/3$ | $\sqrt{6}/3$ |
| 10 | $1/3$ | $2\sqrt{2}/3$ | $-1/3$ | $2\sqrt{2}/3$ | $\sqrt{6}/3$ | $-1/3$ | $2\sqrt{2}/3$ | $-\sqrt{6}/3$ |
| 11 | $1/3$ | $2\sqrt{2}/3$ | $-1/3$ | $2\sqrt{2}/3$ | $-\sqrt{6}/3$ | $1/3$ | $-2\sqrt{2}/3$ | $-\sqrt{6}/3$ |
| 12 | $1/3$ | $2\sqrt{2}/3$ | $-1/3$ | $2\sqrt{2}/3$ | $-\sqrt{6}/3$ | $-1/3$ | $2\sqrt{2}/3$ | $\sqrt{6}/3$ |
| 13 | $1/3$ | $2\sqrt{2}/3$ | $1/3$ | $-2\sqrt{2}/3$ | $\sqrt{6}/3$ | $1/3$ | $-2\sqrt{2}/3$ | $-\sqrt{6}/3$ |
| 14 | $1/3$ | $2\sqrt{2}/3$ | $1/3$ | $-2\sqrt{2}/3$ | $\sqrt{6}/3$ | $-1/3$ | $2\sqrt{2}/3$ | $\sqrt{6}/3$ |
| 15 | $1/3$ | $2\sqrt{2}/3$ | $1/3$ | $-2\sqrt{2}/3$ | $-\sqrt{6}/3$ | $1/3$ | $-2\sqrt{2}/3$ | $\sqrt{6}/3$ |
| 16 | $1/3$ | $2\sqrt{2}/3$ | $1/3$ | $-2\sqrt{2}/3$ | $-\sqrt{6}/3$ | $-1/3$ | $2\sqrt{2}/3$ | $-\sqrt{6}/3$ |
| 17 | $-1/3$ | $-2\sqrt{2}/3$ | $-1/3$ | $2\sqrt{2}/3$ | $\sqrt{6}/3$ | $1/3$ | $-2\sqrt{2}/3$ | $\sqrt{6}/3$ |
| 18 | $-1/3$ | $-2\sqrt{2}/3$ | $-1/3$ | $2\sqrt{2}/3$ | $\sqrt{6}/3$ | $-1/3$ | $2\sqrt{2}/3$ | $-\sqrt{6}/3$ |
| 19 | $-1/3$ | $-2\sqrt{2}/3$ | $-1/3$ | $2\sqrt{2}/3$ | $-\sqrt{6}/3$ | $-1/3$ | $2\sqrt{2}/3$ | $\sqrt{6}/3$ |
| 20 | $-1/3$ | $-2\sqrt{2}/3$ | $-1/3$ | $2\sqrt{2}/3$ | $-\sqrt{6}/3$ | $1/3$ | $-2\sqrt{2}/3$ | $-\sqrt{6}/3$ |
| 21 | $-1/3$ | $-2\sqrt{2}/3$ | $1/3$ | $-2\sqrt{2}/3$ | $\sqrt{6}/3$ | $1/3$ | $-2\sqrt{2}/3$ | $-\sqrt{6}/3$ |
| 22 | $-1/3$ | $-2\sqrt{2}/3$ | $1/3$ | $-2\sqrt{2}/3$ | $\sqrt{6}/3$ | $-1/3$ | $2\sqrt{2}/3$ | $\sqrt{6}/3$ |
| 23 | $-1/3$ | $-2\sqrt{2}/3$ | $1/3$ | $-2\sqrt{2}/3$ | $-\sqrt{6}/3$ | $-1/3$ | $2\sqrt{2}/3$ | $-\sqrt{6}/3$ |
| 24 | $-1/3$ | $-2\sqrt{2}/3$ | $1/3$ | $-2\sqrt{2}/3$ | $-\sqrt{6}/3$ | $1/3$ | $-2\sqrt{2}/3$ | $\sqrt{6}/3$ |
| 25 | $-1/3$ | $2\sqrt{2}/3$ | $-1/3$ | $-2\sqrt{2}/3$ | $\sqrt{6}/3$ | $1/3$ | $2\sqrt{2}/3$ | $-\sqrt{6}/3$ |
| 26 | $-1/3$ | $2\sqrt{2}/3$ | $-1/3$ | $-2\sqrt{2}/3$ | $\sqrt{6}/3$ | $-1/3$ | $-2\sqrt{2}/3$ | $\sqrt{6}/3$ |
| 27 | $-1/3$ | $2\sqrt{2}/3$ | $-1/3$ | $-2\sqrt{2}/3$ | $-\sqrt{6}/3$ | $-1/3$ | $-2\sqrt{2}/3$ | $-\sqrt{6}/3$ |
| 28 | $-1/3$ | $2\sqrt{2}/3$ | $-1/3$ | $-2\sqrt{2}/3$ | $-\sqrt{6}/3$ | $1/3$ | $2\sqrt{2}/3$ | $\sqrt{6}/3$ |
| 29 | $-1/3$ | $2\sqrt{2}/3$ | $1/3$ | $2\sqrt{2}/3$ | $\sqrt{6}/3$ | $-1/3$ | $-2\sqrt{2}/3$ | $-\sqrt{6}/3$ |
| 30 | $-1/3$ | $2\sqrt{2}/3$ | $1/3$ | $2\sqrt{2}/3$ | $\sqrt{6}/3$ | $1/3$ | $2\sqrt{2}/3$ | $\sqrt{6}/3$ |
| 31 | $-1/3$ | $2\sqrt{2}/3$ | $1/3$ | $2\sqrt{2}/3$ | $-\sqrt{6}/3$ | $-1/3$ | $-2\sqrt{2}/3$ | $\sqrt{6}/3$ |
| 32 | $-1/3$ | $2\sqrt{2}/3$ | $1/3$ | $2\sqrt{2}/3$ | $-\sqrt{6}/3$ | $1/3$ | $2\sqrt{2}/3$ | $-\sqrt{6}/3$ |

**Table 2 The 32 isotropic sets of four points on the unit sphere**

Note that the fundamental isotropic set of points given in eq.(2a) is solution 18 of Table 2. Table 3 records seven isotropic sets of points obtained by means of antipodal exchanges.

|            | $P_2$ | $P_3$ | $P_4$ | $P_2\,P_3$ | $P_2\,P_4$ | $P_3\,P_4$ | $P_2\,P_3\,P_4$ |
|------------|-------|-------|-------|------------|------------|------------|------------------|
| Solution # | 10    | 23    | 17    | 16         | 24         | 9          | 15               |

**Table 3 Isotropic sets of points obtained by antipodal exchanges the set #18**

With the forgoing eight isotropic sets of points, we obtain additional isotropic sets by reflections onto the coordinate planes $x$-$z$, $x$-$y$, $x$-$z$ and $x$-$y$; we can then verify that these solutions are listed in Table 2. The corresponding solutions are given in Table 4.

| Solution #                         | 18 | 10 | 23 | 17 | 16 | 24 | 9  | 15 |
|------------------------------------|----|----|----|----|----|----|----|----|
| Reflection plane: $x$-$y$          | 19 | 12 | 22 | 20 | 14 | 21 | 11 | 13 |
| Reflection plane: $x$-$z$          | 27 | 2  | 29 | 28 | 8  | 30 | 1  | 7  |
| Reflection planes: $x$-$z$ and $x$-$y$ | 26 | 4  | 31 | 25 | 6  | 32 | 3  | 5  |

**Table 4 Isotropic sets of points defined by reflections onto the coordinate planes $x$-$y$ and $x$-$z$**

It is noteworthy that Table 4 includes a reflection about the $x$-$y$ and the $x$-$z$ planes, which amount to a 180° rotation about the $x$-axis, and does not lead to a new wrist.

Now, for all seven solutions of Table 3 and the fundamental isotropic set of points given in eq.(2a), we compute the corresponding Denavit-Hartenberg (DH) parameters yielding isotropic wrists. For each isotropic set of points, we place the first joint axis at $P_1$. It is now apparent that we can derive six kinematic chains. Thus, we find $\alpha_1$, $\alpha_2$ and $\alpha_3$ as the angles made by the neighboring position vectors of points $P_i$. Moreover, we eliminate the set of DH parameters leading to wrists that are identical. Hence, a total of eight distinct isotropic wrists are obtained from these sets. The Denavit-Hartenberg parameters of the eight distinct wrists are displayed in Table 5. The corresponding wrists, at their isotropic postures, being displayed in Figs. 2a-h.

| $i$ | $\alpha_i$ | $\theta_i$        |
|-----|------------|-------------------|
| 1   | 109.5°     | $\theta_1$        |
| 2   | 109.5°     | $\pm 60°$         |
| 3   | 109.5°     | $\pm(-60°)$       |
| 4   | *          | $\theta_4$        |

(a)

| $i$ | $\alpha_i$ | $\theta_i$        |
|-----|------------|-------------------|
| 1   | 70.5°      | $\theta_1$        |
| 2   | 109.5°     | $\pm 120°$        |
| 3   | 109.5°     | $\pm 60°$         |
| 4   | *          | $\theta_4$        |

(b)

| $i$ | $\alpha_i$ | $\theta_i$        |
|-----|------------|-------------------|
| 1   | 109.5°     | $\theta_1$        |
| 2   | 70.5°      | $\pm 120°$        |
| 3   | 109.5°     | $\pm 120°$        |
| 4   | *          | $\theta_4$        |

(c)

| $i$ | $\alpha_i$ | $\theta_i$        |
|-----|------------|-------------------|
| 1   | 109.5°     | $\theta_1$        |
| 2   | 109.5°     | $\pm 60°$         |
| 3   | 70.5°      | $\pm 120°$        |
| 4   | *          | $\theta_4$        |

(d)

| $i$ | $\alpha_i$ | $\theta_i$        |
|-----|------------|-------------------|
| 1   | 70.5°°     | $\theta_1$        |
| 2   | 70.5°°     | $\pm 60°$         |
| 3   | 70.5°      | $\pm 60°$         |
| 4   | *          | $\theta_4$        |

(e)

| $i$ | $\alpha_i$ | $\theta_i$        |
|-----|------------|-------------------|
| 1   | 70.5°      | $\theta_1$        |
| 2   | 70.5°      | $\pm 120°$        |
| 3   | 109.5°     | $\pm(-120°)$      |
| 4   | *          | $\theta_4$        |

(f)

| $i$ | $\alpha_i$ | $\theta_i$        |
|-----|------------|-------------------|
| 1   | 109.5°     | $\theta_1$        |
| 2   | 70.5°      | $\pm 120°$        |
| 3   | 70.5°      | $\pm(-60°)$       |
| 4   | *          | $\theta_4$        |

(g)

| $i$ | $\alpha_i$ | $\theta_i$        |
|-----|------------|-------------------|
| 1   | 70.5°      | $\theta_1$        |
| 2   | 109.5°     | $\pm 120°$        |
| 3   | 70.5°      | $\pm(-120°)$      |
| 4   | *          | $\theta_4$        |

(h)

**Table 5 The Denavit-Hartenberg parameters of the eight spherical wrists at their isotropic postures**

Note that the entry corresponding to $\alpha_4$ in the foregoing table is left with an asterisk because this twist angle is not defined for a four-revolute wrist. Its value depends on how the $z$-axis of the task frame is defined. As well, angles $\theta_1$ and $\theta_4$ are left unspecified because isotropy is independent of these values, i.e., isotropy is preserved upon varying these two angles throughout their whole range of values, from $0$ to $2\pi$.

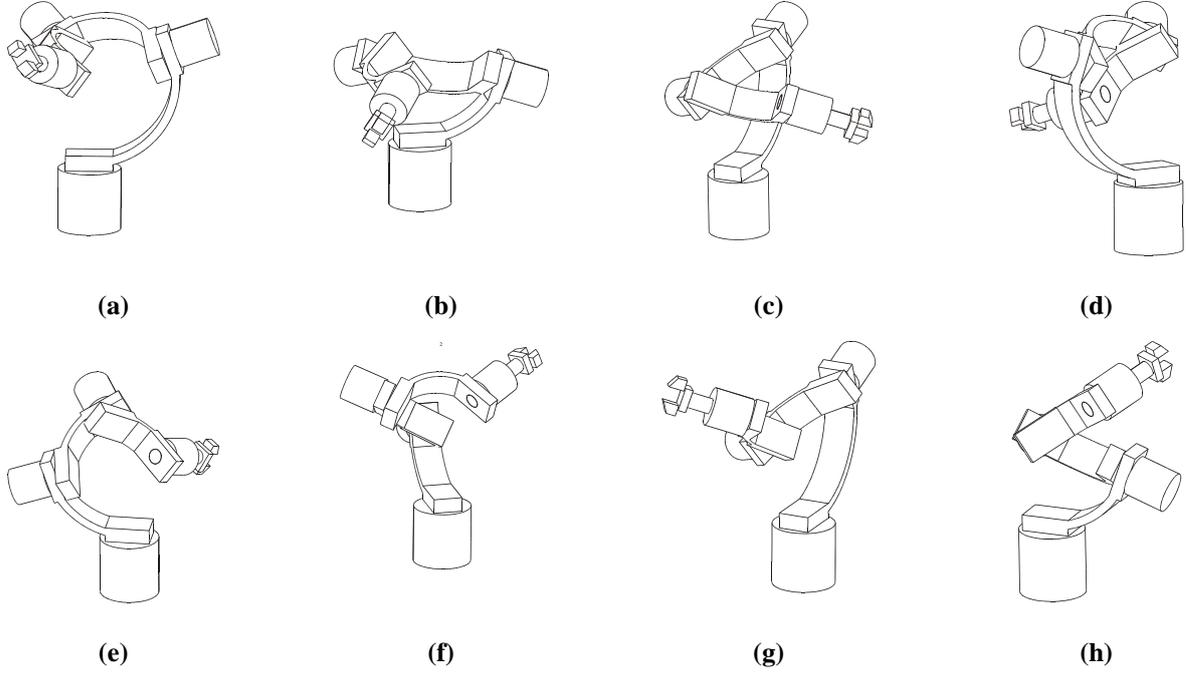

|     (a)     |     (b)     |     (c)     |     (d)     |
|     (e)     |     (f)     |     (g)     |     (h)     |

**Figure 2 The eight distinct isotropic wrists of Table 5**

## 6    Conclusions

We showed that the algebraic formulation of the problem underlying the determination of to all four-revolute serial spherical wrists with kinetostatic isotropy leads to a system of eight quadratic equations in eight unknowns, whose Bezout number is 256, its BKK bound being 192. Nevertheless, this system admits only 32 distinct solutions. Furthermore, upon elimination of the solutions leading to repeated wrists, we are left with only eight distinct isotropic wrists, whose Denavit-Hartenberg parameters were computed, the corresponding wrists having been displayed at their isotropic postures.

## 7    Appendix

We show here that a reflection $\mathbf{L}$ about a line $L$ that passes through the origin is a rotation about $L$ through an angle $\pi$. To this end, we resort to Fig. 3, showing $L$ and a point $P$. To simplify matters, we sketch $P$ and $L$ in their plane.

The *projection* of $P$ onto $L$ is denoted by $P'$, the reflection sought by $P''$, the corresponding position vectors being denoted by $\mathbf{p}$, $\mathbf{p}'$ and $\mathbf{p}''$. Apparently,

$$\mathbf{p}' = \mathbf{e}\mathbf{e}^T\,\mathbf{p} \qquad\qquad (16)$$

Hence,

$$\mathbf{p}'' = \mathbf{p}' - (\mathbf{p} - \mathbf{e}\mathbf{e}^T\mathbf{p}) = (2\mathbf{e}\mathbf{e}^T - \mathbf{1})\,\mathbf{p}$$

Thus, the reflection sought, $\mathbf{L}$, is given by the matrix coefficient of $\mathbf{p}$ in the rightmost side of the foregoing equation, *i.e.*,

$$\mathbf{L} = 2\mathbf{e}\mathbf{e}^T - \mathbf{1} \qquad (17)$$

As the reader can readily verify, the above expression yields a proper orthogonal matrix, and hence, a rotation. Moreover, the axis of the rotation is $\mathbf{e}$ and the angle $\pi$.

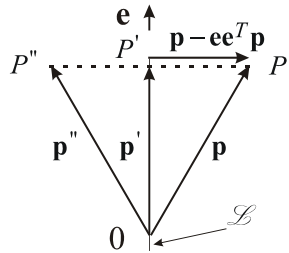

**Figure 3 The reflection of a point with respect to a line**

# 8    References

[1] Salisbury, J. K. and Craig, J. J., "Articulated Hands: Force Control and Kinematic Issues," The Int. J. Robotics Research, Vol. 1, No. 1, pp. 4-17, 1982.

[2] Golub, G. H. and Van Loan, C. F., *Matrix Computations*, The Johns Hopkins University Press, Baltimore, 1989.

[3] Tchoń, K., "Singulrities of Euler wrist, " Mechanism and Machine Theory, Vol. 35, pp. 505-515, 2000.

[4] Long, G. L., Paul, R. P. and Fischer, W. D., "The Hamilton Wrist: A Four-Revolute Spherical Wrist Whiteout Singularities," Proc. IEEE Int. Conf. Robotics and Automation, pp. 902-907, 1989.

[5] Farhang, K. and Zargar, Y. S., "Design of Spherical 4R Mechanisms: Function Generation for the Entire Motion Cycle," ASME J. of Mechanical Design, Vol. 121, pp. 521-528, 1999.

[6] Klein, C. A., and Miklos, T. A., "Spatial Robotic Isotropy," Int. Journal of Robotics Research, Vol. 10, No. 4, August 1991.

[7] Angeles, J., *Fundamentals of Robotic Mechanical Systems*, Second Edition, Springer-Verlag, New York, 2002.

[8] Kroto, H.W., Heath, J.R., O'Brien, S. C., Curl, R.F and Smalley, R.E., "C60: Buckminsterfullerene," Nature, 318, pp. 162-163, 1985.

[9] Angeles, J. and Chablat, D., "On Isotropic Sets of Points in the Plane. Application to the Design of Robot Architectures," in Lenarcic, J. and Stanisic, M. M. (editors), Advances in Robot Kinematic, Kluwer Academic Publishers, pp. 73-82, 2000.

[10] Emiris, I., "Sparse Elimination and Applications in Kinematics," Ph.D. Thesis, UC Berkeley, 1984.

[11] Bernstein, D. N., "The number of roots of a system of equations," Function Analysis and Application, Vol. 9(2), pp. 183-185, 1975.